\documentclass[runningheads]{llncs}
\usepackage{graphicx}
\usepackage[utf8]{inputenc} 
\usepackage[T1]{fontenc}    
\usepackage{hyperref}       
\usepackage{url}            
\usepackage{booktabs}       
\usepackage{amsfonts}       
\usepackage{nicefrac}       
\usepackage{microtype}      
\usepackage{subcaption}
\usepackage{amsmath}
\usepackage{amssymb}
\usepackage{cases}
\usepackage{mathtools}
\usepackage{dirtytalk}
\usepackage{interval}
\usepackage{bm}
\usepackage[ruled,vlined]{algorithm2e}

\usepackage{xcolor}
\begin{document}
\title{FlipOut: Uncovering Redundant Weights via Sign Flipping\thanks{Supported by BrainCreators B.V.}}
%
%
\author{Andrei C. Apostol\inst{1,2,3} \and
Maarten C. Stol\inst{2} \and
Patrick Forr{\'e}\inst{1}}
\authorrunning{Apostol et al.}
%

\institute{Informatics Institute, University of Amsterdam, The Netherlands \and 
BrainCreators B.V., Amsterdam, The Netherlands
\and
\email{apostol.andrei@braincreators.com} }
\maketitle              
\begin{abstract}
Modern neural networks, although achieving state-of-the-art results on many tasks, tend to have a large number of parameters, which increases training time and resource usage. This problem can be alleviated by pruning. Existing methods, however, often require extensive parameter tuning or multiple cycles of pruning and retraining to convergence in order to obtain a favorable accuracy-sparsity trade-off. To address these issues, we propose a novel pruning method which uses the oscillations around $0$ (i.e. sign flips) that a weight has undergone during training in order to determine its saliency. Our method can perform pruning before the network has converged, requires little tuning effort due to having good default values for its hyperparameters, and can directly target the level of sparsity desired by the user. Our experiments, performed on a variety of object classification architectures, show that it is competitive with existing methods and achieves state-of-the-art performance for levels of sparsity of $99.6\%$ and above for most of the architectures tested. For reproducibility, we release our code publicly at \url{https://github.com/AndreiXYZ/flipout}.

\end{abstract}

\section{Introduction}
\label{section:introduction}
The success of deep learning is motivated by competitive results on a wide range of tasks (\cite{attention_is_all_you_need,biggan,densenet}). However, well-performing neural networks often come with the drawback of a large number of parameters, which increases the computational and memory requirements for training and inference. This poses a challenge for deployment on embedded devices, which are often resource-constrained, as well as for use in time sensitive applications, such as autonomous driving or crowd monitoring. Moreover, costs and carbon dioxide emissions associated with training these large networks have reached alarming rates (\cite{carbon_footprint}). To this end, pruning has been proven as an effective way of making neural networks run more efficiently (\cite{obd,obs1,pruning_filters_for_efficient_convnet,han_magnitude,molchanov_vd}). 

Early works (\cite{obs1,obd}) have focused on using the second-order derivative to detect which weights to remove with minimal impact on performance. However, these methods either require strong assumptions about the properties of the Hessian, which are typically violated in practice, or are intractable to run on modern neural networks due to the computations involved.

One could instead prune the weights whose optimum lies at or close to $0$ anyway. Building on this idea, the authors of  \cite{han_magnitude} propose training a network until convergence, pruning the weights whose magnitudes are below a set threshold, and allowing the network to re-train, a process which can be repeated iteratively. This method is improved on in \cite{lth}, whereby the authors additionally reset the remaining weights to their values at initialization after a pruning step. Yet, these methods require re-training the network until convergence multiple times, which can be a time consuming process.

Recent alternatives either rely on methods typically used for regularization (\cite{L0,deephoyer,molchanov_vd}) or introduce a learnable threshold, below which all weights are pruned (\cite{dynamic_sparse_training}). All these methods, however, require extensive hyperparameter tuning in order to obtain a favorable accuracy-sparsity trade-off. Moreover, the final sparsity of the resulting network cannot be predicted given a particular choice of these hyperparameters. These two issues often translate into the fact that the practitioner has to run these methods multiple times when applying them to novel tasks.

To summarize, we have seen that the pruning methods presented so far suffer from one or more of the following problems: (1) computational intractability, (2) having to train the network to convergence multiple times, (3) requiring extensive hyperparameter tuning for optimal performance and (4) inability to target a specific final sparsity.

We note that by using a heuristic in order to determine during training whether a weight has a locally optimal value of low magnitude, pruning can be performed before the network reaches convergence, unlike the method proposed by the authors of \cite{han_magnitude}. We propose one such heuristic, coined \textit{the aim test}, which determines whether a value represents a local optimum for a weight by monitoring the number of times that weight oscillates around it during training, while also taking into account the distance between the two. We then show that this can be applied to network pruning by applying this test at the value of $0$ for all weights simultaneously, and framing it as a saliency criterion. By design, our method is tractable, allows the user to select a specific level of sparsity and can be applied during training.

Our experiments, conducted on a variety of object classification architectures, indicate that it is competitive with respect to relevant pruning methods from literature, and can outperform them for sparsity levels of $99.6\%$ and above. Moreover, we empirically show that our method has default hyperparameter settings which consistently generate near optimal results, easing the burden of tuning.

\section{Method}
\label{section:method}
\subsection{Motivation}
\label{subsection:motivation}

\begin{figure}[t]
  \centering
  \includegraphics[width=60mm]{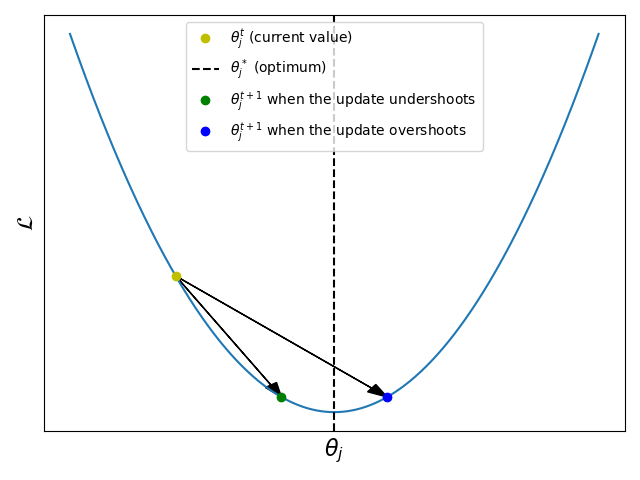}
  \caption{Over- and under-shooting illustrated. The vertical line splits the x-axis into two regions relative to the (locally-)optimal value $\theta_j^*$. Overshooting corresponds to when a weight gets updated such that its new value lies in the opposite region (blue dot), while undershooting occurs when the updated value is closer to the optimal value, but stays in the same region (green dot).}
  \label{fig:over_under_shooting}
\end{figure}

Mini-batch stochastic gradient descent (\cite{bottou_sgd_batch}) is the most commonly used optimization method in machine learning. Given a mini-batch of $B$ randomly sampled training examples consisting of pairs of features and labels $\{(x_b,y_b)\}_{b=1}^B$, a neural network parameterised by a weight vector $\bm{\theta}$, a loss objective $\mathcal{L}(\bm{\theta}, \bm{x}, \bm{y})$ and a learning rate $\eta$, the update rule of stochastic gradient descent is as follows:
\begin{align*}
    \bm{g}^t &= \frac{1}{B} \sum_{b=1}^B \nabla_{\bm{\theta}^t} \mathcal{L}(\bm{\theta}^t, x_b, y_b) \\
    \bm{\theta}^{t+1} &\leftarrow \bm{\theta}^t - \eta \bm{g}^t
\end{align*}
\noindent Given a weight $\theta_j^t$, one could consider its possible values as being split into two regions, with a locally optimal value $\theta_j^*$ as the separation point. Depending on the value of the gradient and the learning rate, the updated weight $\theta_j^{t+1}$ will lie in one of the two regions. That is, it will either get closer to its optimal value while remaining in the same region as before or it will be updated past it and land in the opposite region. We term these two phenomena under- and over-shooting, and provide an illustration in Fig. \ref{fig:over_under_shooting}. Mathematically, they correspond to $\eta|g^t_j| < |\theta^t_j - \theta_j^*|$ and $\eta|g^t_j| > |\theta^t_j - \theta_j^*|$, respectively.

With the behavior of under- and over-shooting, one could construct a heuristic-based test in order to evaluate whether a weight has a local optimum at a specific point without needing the network to have reached convergence: (1) for a weight $\theta_j$, a value of $\phi_j$ is chosen for which the test is conducted,  (2) train the model regularly and record the occurrence of under- and over-shooting around $\phi_j$ after each step of SGD, (3) if the number of such occurrences exceeds a threshold $\kappa$, conclude that $\theta_j$ has a local optimum at $\phi_j$, i.e. $\theta_j^*=\phi_j$. We coin this method \textit{the aim test}. 

Previous works have demonstrated that neural networks can tolerate high levels of sparsity with negligible deterioration in performance (\cite{han_magnitude,molchanov_vd,lth,dynamic_sparse_training}). It is then reasonable to assume that for a large number of weights, there exist local optima at exactly $0$, i.e. $\theta_j^*=0$. One could then use the aim test to detect these weights and prune them. Importantly, when using the aim test for $\phi_j=0$, the two regions around the tested value are the set of negative and positive real numbers, respectively. Checking for over-shooting then becomes equivalent to testing whether the sign of $\theta_j$ has changed after a step of SGD, while under-shooting can be detected when a weight has been updated to a smaller absolute value and retained its sign, i.e. $(|\theta_j^{t+1}| < |\theta_j^{t}|) \wedge (\text{sgn}(\theta_j^t) = \text{sgn}(\theta_j^{t+1}))$.

\begin{figure}[t]
\centering
\begin{subfigure}{.5\textwidth}
  \centering
  \includegraphics[width=60mm]{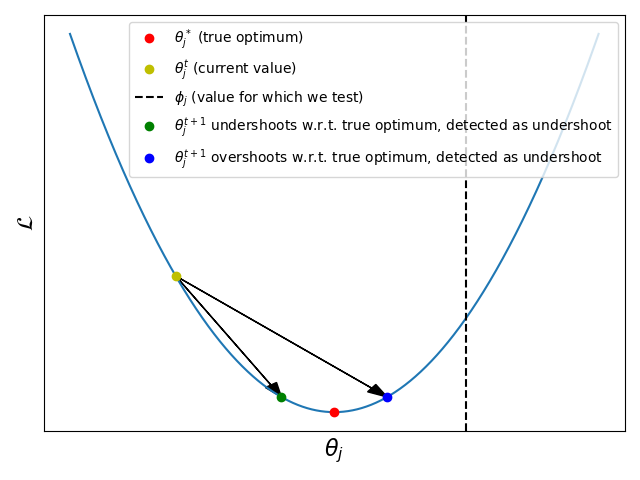}
  \caption{Deceitful observations of under-shooting.}
  \label{fig:weight_is_far}
\end{subfigure}%
\begin{subfigure}{.5\textwidth}
  \centering
  \includegraphics[width=60mm]{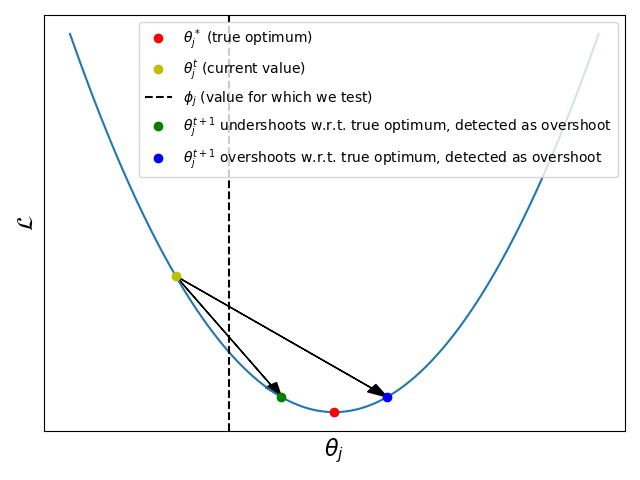}
  \caption{Deceitful observations of over-shooting.}
  \label{fig:overshooting_weight_is_far}
\end{subfigure}
\caption{In the plots above, the dotted vertical line represents the value at which the aim test is conducted (i.e. a value we would like to determine as a local optimum or not), while the red dot represents the value of a true local optimum. When testing for a value which is not a locally optimal value $\phi_j \neq \theta_j^*$, over- or under-shooting around $\phi_j$ can be merely a side-effect of that weight getting updated towards its true optimum $\theta_j^*$.  These observations would then contribute towards the aim test returning a false positive outcome, i.e. $\phi_j = \theta_j^*$. Whether we observe an over-shoot or an under-shoot in this case depends on the relationship between $\phi_j$ and $\theta_j^*$. In (a), we have $\phi_j > \theta_j^*$, where if the hypothesised and true optimum are sufficiently far apart, we observe an under-shoot. Conversely, in (b), we have $\phi_j < \theta_j^*$ and observe over-shooting.}
\end{figure}

However, under-shooting can be problematic; for instance, a weight could be updated to a lower magnitude, while at the same time being far from $0$. This can happen when a weight is approaching a non-zero local optimum, an occurrence which should not contribute towards a positive outcome of the aim test.  By positive outcome, we refer to determining that $\phi_j=0$ is indeed a local optimum of $\theta_j$. A similar problem can occur for over-shooting, where a weight receives a large update that causes it to change its sign but not lie in the vicinity of $0$. These scenarios, which we will refer to as \textit{deceitful shots} going forward, are illustrated in the general case, where $\phi_j$ can take any value, in Fig.  \ref{fig:weight_is_far} and Fig. \ref{fig:overshooting_weight_is_far}. Following, we make two observations which help circumvent this problem. 

Firstly, one could reduce the impact of deceitful shots by also taking into account the distance of the weight to the hypothesised local optimum, i.e. $|\theta_j - \phi_j|$, when conducting the aim test. In other words, the number of occurrences of under- and over-shooting should be weighed inversely proportional to this quantity, even if they would otherwise exceed $\kappa$.

\begin{figure}[t]
\centering
\begin{subfigure}{.5\textwidth}
  \centering
  \includegraphics[width=60mm]{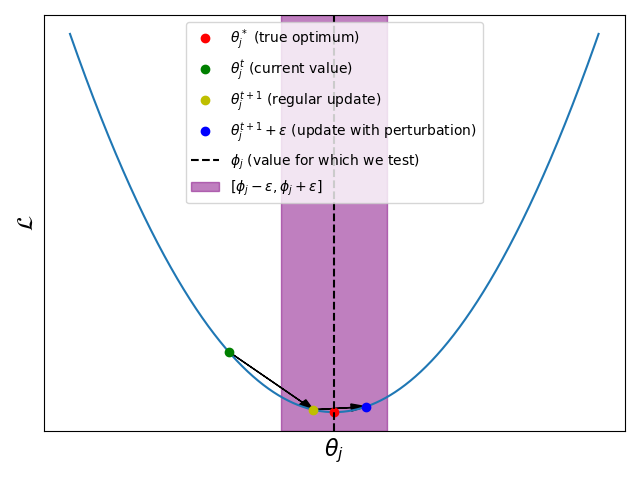}
  \caption{Under-shooting to over-shooting.}
  \label{fig:update_with_noise_overshoots}
\end{subfigure}%
\begin{subfigure}{.5\textwidth}
  \centering
  \includegraphics[width=60mm]{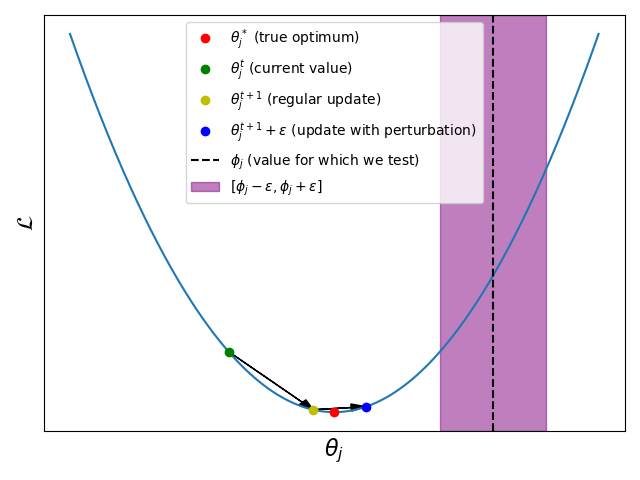}
  \caption{Ignoring deceitful shots.}
  \label{fig:undershooting_not_considered}
\end{subfigure}
\caption{(a) All weights that under-shoot but are within $\epsilon$ of $\phi_j$ will be made to over-shoot. (b) When testing at a value which is not a local optimum for $\theta_j$, i.e. $\phi_j \neq \theta_j^*$ and adding a perturbation $\epsilon$ to $\theta_j$, not taking under-shooting into account means that if the weight gets updated such that it does not lie in the boundary around $\phi_j$ induced by the perturbation, an event that would otherwise contribute to a false positive outcome for the aim test will not be recorded, so the likelihood of rejecting $\phi_j$ as an optimum increases.}
\end{figure}

Our second observation is that by ignoring updates which are not in the vicinity of $\phi_j$, the number of deceitful shots are reduced. In doing so, one could also simplify the aim test; with a sufficiently large perturbation to $\theta_j$, an update that might otherwise cause under-shooting can be made to cause over-shooting. Adding a perturbation of $\pm \epsilon$ is, in effect, inducing a boundary around the tested value, $\interval[scaled]{\phi_j-\epsilon}{\phi_j+\epsilon}$; all weights that get updated such that they fall into that boundary will be said to over-shoot around $\phi_j$. With this framework, checking for over-shooting is sufficient; updates that under-shoot and are within $\epsilon$ of the tested value are made to over-shoot (Fig. \ref{fig:update_with_noise_overshoots}) and updates which under-shoot but are not in the vicinity of $\phi_j$, i.e. a deceitful shot, are now not recorded at all (Fig. \ref{fig:undershooting_not_considered}). This can also be seen as restricting the aim test to only operate within a vicinity around $\phi_j$.

\subsection{FlipOut: applying the aim test for pruning}
\label{section:flipout_applying_the_aim_test}
\subsubsection{Determining which weights to prune}
\label{section:saliency}
Pruning weights that have local optima at or around $0$ can obtain a high level of sparsity with minimal degradation in accuracy. The authors of \cite{han_magnitude} use the magnitude of the weights once the network is converged as a criterion; that is, the weights with the lowest absolute value (i.e. closest to $0$) get pruned. The aim test can be used to detect whether a point represents a local optimum for a weight and can be applied before the network reaches convergence, during training. For pruning, one could then apply the aim test simultaneously for all weights with $\bm{\phi} = \bm{0}$ . We propose framing this as a saliency score; at time step $t$, the saliency $\tau_j^t$ of a weight $\theta_j^t$ is:
\begin{subequations}
\label{eqn:flipout_eqns}
\begin{align}
    \tau_j^t &= \frac{|\theta_j^t|^{p}}{\text{flips}_j^t} \label{eqn:flipout} \\
    \text{flips}_j^t &= \sum_{i=0}^{t-1} [\text{sgn}(\theta_j^i) \neq \text{sgn}(\theta_j^{i+1})] \label{eqn:flips}
\end{align}
\end{subequations}
With perturbation added into the weight vector, it is enough to check for over-shooting, which is equivalent to counting the number of sign flips a weight has undergone during the training process when $\phi_j = 0$ (Eq. \ref{eqn:flips}); a scheme for adding such perturbation is described in Section \ref{section:gradnoise}. In Equation \ref{eqn:flipout}, the denominator $|\theta_j^t|^p$ represents the proximity of the weight to the hypothesised local optimum, $|\theta_j^t - \phi_j|^p$ (which is equivalent to the weight's magnitude since we have $\phi_j=0$ for all weights). The hyperparameter $p$ controls how much this quantity is weighted relative to the number of sign flips.

When determining the amount of parameters to be pruned, we adopt the strategy from \cite{lth}, i.e. pruning a percentage of the remaining weights each time, which allows us to target an exact level of sparsity. Given $m$, the number of times pruning is performed, $r$ the percentage of remaining weights which are removed at each pruning step, $k$ the total number of training steps, $d_\theta$ the dimensionality of the weights and $||\cdot||_0$ the $L_0$-norm, the resulting sparsity $s$ of the weight tensor after training the network is simply:
\begin{equation}
    s = 1 - \frac{||\bm{\theta}^k||_0}{d_{\bm{\theta}}} = (1-r)^{m}
\end{equation}
This final sparsity can then be determined by setting $m$ and $r$ appropriately.

\subsubsection{Perturbation through gradient noise}
\label{section:gradnoise}

Adding gradient noise has been shown to be effective for optimization (\cite{adding_gradient_noise,welling_sgld}) in that it can help lower the training loss and reduce overfitting by encouraging an exploration in the parameter space, thus effectively acting as a regularizer.
While the benefits of this method are helpful, our motivation for its usage stems from allowing the aim test to be performed in a simpler manner; weights that get updated closer to $0$ will occasionally pass over the axis due to the injected noise, thus making checking for over-shooting sufficient. We scale the variance of the noise distribution by the $L_2$ norm of the parameters $\bm{\theta}$, normalize it by the number of weights and introduce a hyperparameter $\lambda$ which scales the amount of noise added into the gradients. For a layer $l$ and $d_l$ its dimensionality, the gradient for the weights in that layer used by SGD for updates will be:
\begin{subequations}
\label{eqn:noise}
\begin{align}
    \bm{\hat{g}}^{t,l} &\leftarrow \bm{g}^{t,l} + \lambda \bm{\epsilon}^{t,l}  \label{eqn:noise_addition}\\
    \bm{\epsilon}^{t,l} &\sim \mathcal{N}(0, \sigma_{t,l}^2) \\
    \sigma_{t,l}^2 &= \frac{\| \bm{\theta}^{t,l}\|_2^2}{d_l} \label{variance_scale}
\end{align}
\end{subequations}
As training is performed, it is desirable to reduce the amount of added noise so that the network can successfully converge. Previous works use annealing schedules by decaying the variance of the Gaussian distribution  proportional to the current time step. Under our proposed formulation, however, explicitly using an annealing schema is not necessary. By pruning weights, the term in the numerator in Eq. \ref{variance_scale} decreases, while the denominator remains constant. This ensures that annealing will be induced automatically through the pruning process, and there is no need for manually constructing a schedule. 

Pruning periodically throughout training according to the saliency score in Eq. \ref{eqn:flipout} in conjunction with adding gradient noise into the weights forms the \textit{FlipOut} pruning method.

\section{Related work}
\subsection{Deep-R}
In Deep-R (\cite{deepr}), the authors split the weights of the neural network into two matrices, the connection parameter $\theta_k$ and a constant sign $s_k$ with $s_k \in \{-1, +1\}$; the final weights of the network are then defined as $\bm{\theta} \odot \bm{s}$. The connections whose $\theta_k$ is negative are inactive; whenever a connection changes its sign, it is turned dormant and another randomly sampled connection is re-activated, ensuring the same sparsity level is maintained throughout training. Gaussian noise is also injected into the gradients during training.

Two similarities with our method can be observed here, namely the fact that the authors also use sign flipping as a signal for pruning a weight, and the addition of Gaussian noise. However, our methods differ in that we do not impose a set level of sparsity throughout training; instead, we use the number of sign flips of a weight in order to determine its saliency, while in Deep-R a single sign flip is required for a weight to be removed. 
Our method of injecting noise into the gradients also differs in that it does not explicitly encode an annealing scheme, allowing for the pruning process itself to reduce the noise throughout training.

\subsection{Magnitude and uncertainty pruning}
The M\&U pruning criterion is proposed in \cite{weight_and_uncertainty_pruning}. Given a weight $\theta_j$, its uncertainty estimate $\Tilde{\sigma}_{\theta_j}$ and a parameter $\lambda$ controlling the trade-off between magnitude and uncertainty, the M\&U criterion will evaluate the saliency of the weight as:
\begin{align*}
    \tau_j = \frac{|\theta_j|}{\lambda + \Tilde{\sigma}_{\theta_j}}
\end{align*}
Uncertainty is estimated as the standard deviation across the previous $n$ values of that weight, via a process called  pseudo-bootstraping.
This criterion is a generalization of the Wald test, and is equivalent to it when $\lambda=0$. 

Our method is similar in that our saliency score also normalizes the weight's magnitude by a function of its past values. However, this method assumes asymptotic normality. While this is the case when using negative log-likelihood or an equivalent as the loss function, this property does not necessarily hold when using modified variants of the SGD estimator, such as Adam (\cite{adam}) or RMSprop (\cite{rmsprop}). In contrast, FlipOut is not derived from the Wald test and does not make any assumptions about the weight distribution at convergence.

\section{Experiments}
\label{section:experiments}
\subsection{General Setup}
\label{section:experimental_setup}
\subsubsection{Baselines}
As baselines, we consider a slightly modified version of magnitude pruning (\cite{han_magnitude}) (Global magnitude), due to the similarity between its saliency criterion and that of our own method, SNIP (\cite{snip}) due to it being an easily applicable method which does not suffer from any of the issues that are commonly found in pruning methods (Section \ref{section:introduction}) and Hoyer-Square, as introduced in \cite{deephoyer}, for the state-of-the-art results that it has demonstrated. We also include random pruning (Random) as a control. For FlipOut, Global magnitude and Random, pruning is performed periodically throughout training. We compare these methods at five different compression ratios, chosen at regular log-intervals (Table \ref{table:sparsity_levels}); for Hoyer-square, the performance at those points is estimated by a sparsity-accuracy trade-off curve. Magnitude pruning, in its original formulation, performs pruning only once the network has reached convergence. However, employing this strategy can create a confounding variable: training time. Since we would like to compare all methods at equal training budgets, we have opted to simply perform pruning after a fixed number of epochs for these methods. Note that the training budget that we allocate allows all of the networks that we consider to reach convergence when trained without performing any pruning. We make an exception to this equal budget rule for Hoyer-Square, since it prunes after training and would otherwise not benefit from any SGD updates after sparsification. As such, we have performed an additional $150$ epochs of fine-tuning without the regularizer, as per the original method, although we have observed negligible benefits to this. All baselines were modified to rank the weights globally when a pruning decision is made, as per the strategy from \cite{lth}, in order to avoid creating bottleneck layers. The models that we test on are ResNet18 (\cite{resnet}) and VGG19 (\cite{vgg}) trained on the CIFAR-10 dataset (\cite{cifar10}) and DenseNet121 (\cite{densenet}) trained on Imagenette (\cite{imagenette_dataset}). 
\begin{table}
  \caption{Compression ratios, resulting sparsity levels and prune frequencies used in the experiments, assuming $350$ epochs of training and that $50\%$ of the remaining weights are removed at each step.}
  \label{table:sparsity_levels}
  \centering
  \begin{tabular}{lll}
    \toprule
    Compression ratio ($\frac{d_\theta}{||\theta||_{0}}$) & Resulting sparsity ($1-\frac{||\theta||_{0}}{d_\theta}$) & Epochs before pruning \\
    \midrule
    $2^2$ & $75\%$ & $117$\\
    $2^4$ & $93.75\%$ & $70$ \\
    $2^6$ & $98.43\%$ & $50$ \\
    $2^8$ & $99.61\%$ & $39$ \\
    $2^{10}$ & $99.9\%$ & $32$ \\
    \bottomrule
  \end{tabular}
\end{table}
\subsubsection{Hyperparameters} The training parameters for all experiments are taken from \cite{pytorch_cifar_10}; specifically, we use a learning rate of $0.1$, batch size of $128$, $350$ epochs of training and a weight decay penalty of $5e-4$. The learning rate is decayed by a factor of $10$ at epochs $150$ and $250$. The networks are trained with the SGD optimizer with a momentum value of $0.9$ (\cite{bottou_sgd_batch}). For the methods that perform iterative pruning (Global magnitude, Random, FlipOut), we remove $50\%$ of the remaining weights at each pruning step, with the pruning frequencies chosen such that the compression ratios from Table \ref{table:sparsity_levels} are achieved; we use the same pruning rates and frequencies across all three methods. SNIP accepts a single hyperparameter, namely the desired final sparsity, which we have chosen such that it matches the aforementioned compression ratios. For Hoyer-Square, which does not allow for a specific level of sparsity to be chosen and, instead, relies on parameter tuning, we generate a sparsity-accuracy trade-off curve by using $15$ different values for the regularization term, ranging from $1e-7$ to $6e-3$ with $3$ values at each decimal point (e.g. $1e-7$, $3e-7$, $6e-7$, $1e-6$ etc.)  and a fixed pruning threshold of $1e-4$. Finally, for FlipOut, we use the values of $p=2$ (Eq. \ref{eqn:flipout_eqns}) and $\lambda=1$ (Eq. \ref{eqn:noise}) for all experiments, a choice we motivate in Section \ref{section:choosing_hyperparameters}.

\subsection{Choosing the hyperparameters for FlipOut}
\label{section:choosing_hyperparameters}
We have experimented with different values of the two hyperparameters and found that $p=2$ (Eq. \ref{eqn:flipout}) and $\lambda=1$ (Eq. \ref{eqn:noise_addition}) offer consistent, strong results for all networks tested. In the following paragraphs, we detail the procedure used in determining these values.

\subsubsection{Choosing $\bm{\lambda}$} For $\lambda$, we have run all networks at $15$ different values, ranging from $0.75$ to $1.5$ in increments of $0.05$. The value of $p=2$ was used. The networks are evaluated on a validation set, created by removing a random subset of samples from the training set. The size of the validation set was $10000$ for CIFAR10 and $2000$ for Imagenette. For our subsequent experiments, (Sections \ref{section:results} and \ref{section:noise_ablation}), the networks have been trained on the full training set.  As a metric, we have used the accuracy of the networks at the end of training for the sparsity levels of $93.75\%$ and $99.9\%$. We provide in Table \ref{table:lambda_gridsearch} the accuracies generated by the optimal value of $\lambda$, as discovered through this process, and the ones generated at $\lambda=1$. Notice that the differences are almost negligible at $93.75\%$ sparsity. For the larger sparsity level the disparity increases, although the default value still remains within $2$ percentage points of the optimum value for all networks considered. The largest gap can be seen for ResNet18 and DenseNet121, at approximately $1.7$ and $1.5$ percentage points, respectively. Since there are only two out of six cases in which optimizing $\lambda$ has helped beyond a negligible amount, we have used the value of $1$ for this hyperparameter throughout our experiments.

\begin{table}[t]
  \caption{Accuracies when using the best value of $\lambda$ discovered by grid search and the value of $\lambda=1$ at two levels of sparsity. The parantheses indicate the gain offered by the optimal parameter.}
  \label{table:lambda_gridsearch}
  \centering
  \begin{tabular}{lllll}
    \toprule
    & \multicolumn{2}{c}{Acc. at sparsity $93.75\%$} & \multicolumn{2}{c}{Acc. at sparsity $99.9\%$} \\
    \cmidrule(r){2-3}
    \cmidrule(r){4-5}
    Model & $\lambda^*$ & $\lambda=1$ & $\lambda^*$ & $\lambda=1$ \\
    \midrule
    ResNet18 & 94.58(+0.02) & 94.56 & 83.75(+1.68) & 82.07 \\
    VGG19 & 93.07(+0.11) & 92.96 & 87.72(+0.48) & 87.24 \\
    DenseNet121 & 89.75(+0.0) & 89.75 & 73.5(+1.45) & 72.05 \\
    \bottomrule
  \end{tabular}
\end{table}

\begin{table}[t]
  \caption{Table of results for different values of $p$ at two levels of sparsity.}
  \label{table:p_gridsearch}
  \centering
  \resizebox{\textwidth}{!}{
  \begin{tabular}{lllllllllll}
    \toprule
    & \multicolumn{5}{c}{Acc. at sparsity $93.75\%$} & \multicolumn{5}{c}{Acc. at sparsity $99.9\%$} \\
    \cmidrule(r){2-6}
    \cmidrule(r){7-11}
    Model & $p=0$ & $p=\frac{1}{2}$ & $p=1$ & $p=2$ & $p=4$ & $p=0$ & $p=\frac{1}{2}$ & $p=1$ & $p=2$ & $p=4$ \\
    \midrule
    ResNet18 & 93.71 & 88.39 & 94.18 & \textbf{94.26} & 94.11 & 72.69 & 77.08 & 79.83 & 82.07 & \textbf{83.15} \\
    VGG19 & 91.68 & 82.44 & 92.56 & \textbf{92.96} & 92.57 & 81.48 & 80.69 & 86.01 & \textbf{87.24} & 86.64 \\
    DenseNet121 & 10.35 & 77.40 & 88.9 &  \textbf{89.75} & 88.86 & 10.35 & 10.35 & 70.85 & \textbf{72.05} & 60.55 \\
    \bottomrule
  \end{tabular}}
\end{table}

\subsubsection{Choosing $\bm{p}$} We perform similar experiments for $p$ on five values, $p \in \{0, \frac{1}{2}, 1, 2, 4\}$. Note that the value of $p=0$ corresponds to the case when the magnitudes of the weights are not taken into account; that is, the pruning decisions will be made solely based on the number of sign flips. As can be seen in Table \ref{table:p_gridsearch}, the value of $p=2$ consistently outperforms all other tested values, with the exception of ResNet18 at $99.9\%$ sparsity, for which the value of $p=4$ achieves better results by approximately $1$ percentage point. Another interesting observation is that the values of $1$, $2$ and $4$ tend to perform better than $0$ and $\frac{1}{2}$; we conjecture that this is due to the fact that deceitful shots (Section \ref{subsection:motivation}) occur when not taking into account the distance between the weight and its hypothesised local optimum, which have a negative impact on the pruning decision. This can be especially observed at the higher sparsity level and in the case of DenseNet121, where pruning with $p=0$ causes the network to not perform better than random guessing. Given that the value of $p=2$ is favored in $5$ out of $6$ cases, we have decided to use it as a default value in our subsequent experiments.

\subsection{Comparison to baselines}
\label{section:results}

\begin{figure}[t]
\centering
\begin{subfigure}{.5\textwidth}
  \centering
  \includegraphics[width=60mm]{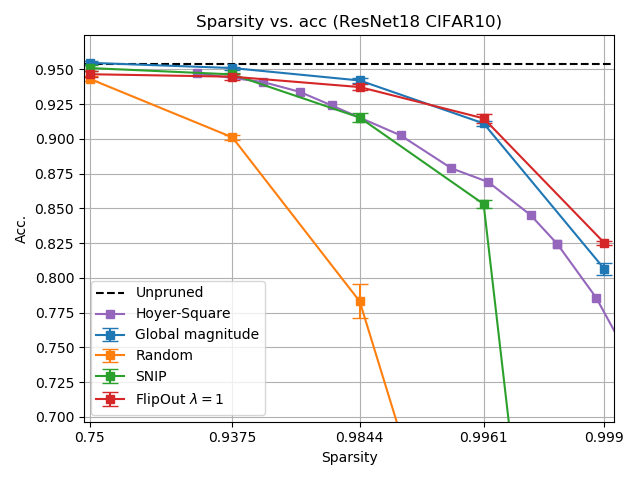}
  \caption{ResNet18 on CIFAR10}
  \label{fig:results_unstructured_rn18}
\end{subfigure}%
\begin{subfigure}{.5\textwidth}
  \centering
  \includegraphics[width=60mm]{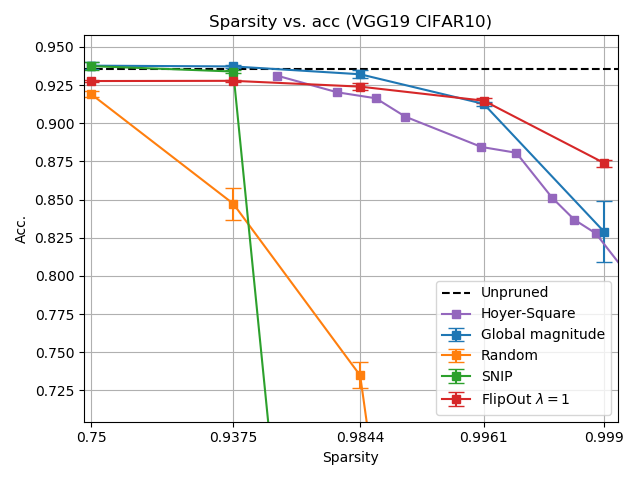}
  \caption{VGG19 on CIFAR10}
  \label{fig:results_unstructured_vgg19}
\end{subfigure}
\\
\begin{subfigure}{.5\textwidth}
  \centering
  \includegraphics[width=60mm]{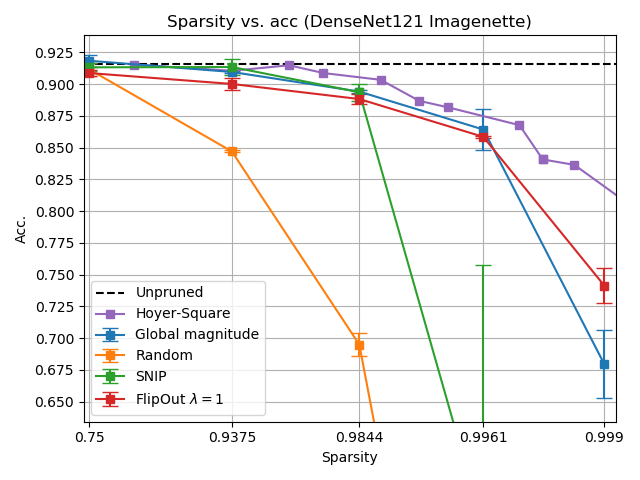}
  \caption{DenseNet121 on Imagenette}
  \label{fig:results_unstructured_dn121}
\end{subfigure}%

\caption{Results of pruning ResNet18 and VGG19 on the CIFAR10 dataset. Each point represents an average over 3 runs with error bars indicating standard deviation. The accuracy of the unpruned model is included for reference.}
\label{fig:results_unstructured}
\end{figure}

The results for the three models tested are found in Figure \ref{fig:results_unstructured}. FlipOut obtains state-of-the-art performance on ResNet18 and VGG19 for sparsity levels of $99.61\%$ and beyond. For the highest tested sparsity level, it outperforms the second-best method by $1.9$ and $4.5$ percentage points, respectively (Fig. \ref{fig:results_unstructured_rn18}, \ref{fig:results_unstructured_vgg19}). Notably, when using FlipOut on VGG19 for this sparsity, the drop in accuracy compared to the unpruned model is only $6.2$ percentage points. At the same time, it remains competitive with other baselines for lower degrees of sparsity, staying within a $1$ percentage point difference compared to the best method and with a minimal drop relative to the unpruned model. For DenseNet121, however, Hoyer-Square dominates all other methods tested in most cases (Fig. \ref{fig:results_unstructured_dn121}), with FlipOut as second best for the highest sparsity level.

Interestingly, the simple criterion of magnitude pruning, when modified to rank the weights globally instead of a layer-by-layer basis, is competitive with other, more recent, baselines, and even obtains state-of-the-art results for moderate levels of sparsity. However, at high levels of sparsity, which correspond to more frequent and implicitly earlier pruning steps (Table \ref{table:sparsity_levels}) there is a performance degradation. This suggests that the magnitude of a weight by itself is not a good measure of saliency when the network is far from reaching convergence. It is also worth noting that SNIP collapses at high levels of sparsity, in some cases performing even worse than random pruning (Fig. \ref{fig:results_unstructured_vgg19}). We conjecture that this behavior is a consequence of pruning at initialization; removing too many parameters too early can impede training. Upon inspecting the cases where SNIP collapses (not shown in figures for visibility purposes) we noticed that at least one layer has been entirely pruned, effectively blocking any signal from passing. Interestingly, this does not happen for any of the other baselines (except for Random).

We note that unlike Hoyer-Square, our method does not require extensive parameter tuning and can target the final sparsity directly. Moreover, FlipOut can be applied during training and does not need additional epochs of fine-tuning. Finally, SNIP, the only other baseline which does not suffer from any of the issues commonly found among pruning methods (Section \ref{section:introduction}) compromises on performance for high levels of sparsity, whereas FlipOut does not.

\subsection{Is it just the noise?}
\label{section:noise_ablation}

\begin{figure}[t]
\centering
\begin{subfigure}{.5\textwidth}
  \centering
  \includegraphics[width=60mm]{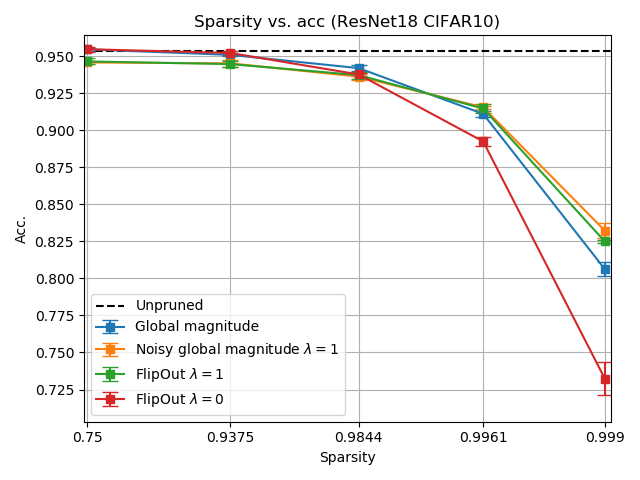}
  \caption{ResNet18 on CIFAR10}
  \label{fig:noise_ablation_rn18}
\end{subfigure}%
\begin{subfigure}{.5\textwidth}
  \centering
  \includegraphics[width=60mm]{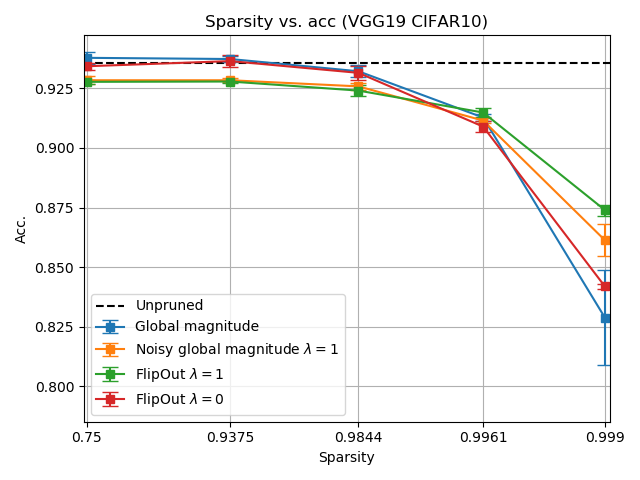}
  \caption{VGG19 on CIFAR10}
  \label{fig:noise_ablation_vgg19}
\end{subfigure}
\\
\begin{subfigure}{.5\textwidth}
  \centering
  \includegraphics[width=60mm]{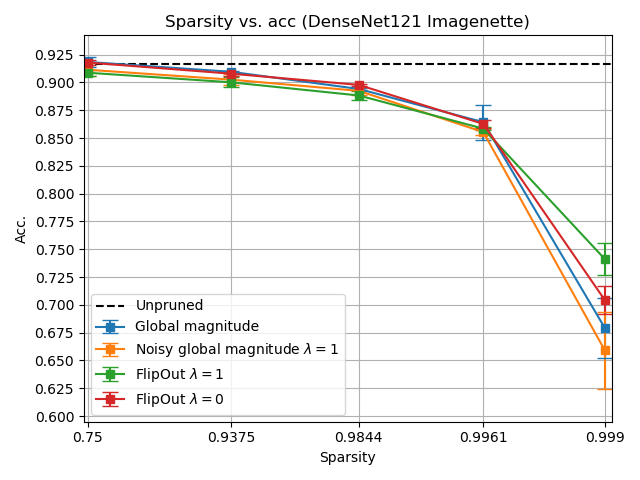}
  \caption{DenseNet121 on Imagenette}
  \label{fig:noise_ablation_dn121}
\end{subfigure}
\caption{Results of the ablation study on the noise. Global magnitude without adding noise is also shown for comparison. Each point is the average over 3 different seeds; error bars indicate standard deviation.}
\label{fig:noise_ablation}
\end{figure}

The performance of FlipOut could simply be a result of the noise addition, which is known to aid optimization (\cite{adding_gradient_noise,welling_sgld}). To investigate this, we perform experiments with global magnitude as the pruning criterion in which we add noise into the gradients using the recipe from Equation $\ref{variance_scale}$ and compare it to our own method. Notably, the saliency criterion of these two methods differ only in that FlipOut normalizes the magnitude by the number of sign flips (denominator in Eq. \ref{eqn:flipout}). The hyperparameters were kept at their default values of $p=2$ for FlipOut and $\lambda=1$ for both methods. We also include runs of FlipOut where no noise was added (i.e. $\lambda=0$). These serve as a control, decoupling the two novel components of our method: noise addition and scaling magnitudes by the number of sign flips.  The same pruning rates and frequency of pruning steps have been used as before (Table \ref{table:sparsity_levels}). The results are illustrated in Fig. \ref{fig:noise_ablation}.

For sparsity levels up to $98.44\%$, adding gradient noise causes a slight deterioration on performance, as can be seen by the fact that both global magnitude and FlipOut with $\lambda=0$ outperform their noisy counterparts. It can also be seen that FlipOut with $\lambda=1$ performs comparably to noisy global magnitude, indicating that measuring saliency by sign flips does not benefit accuracy in these regimes compared to using only the magnitude, and the performance gap between the noisy and non-noisy methods is likely a result of noise addition. For sparsity levels of $99.61\%$ and above, however, the opposite is true. It seems that gradient noise disproportionately benefits networks with a small number of remaining parameters; we conjecture that this is due to the fact that the exploration in parameter space induced by noise is more effective when that space is heavily constrained. 
Focusing on the highest level of sparsity, FlipOut outperforms noisy global magnitude on VGG19 (Fig. \ref{fig:noise_ablation_vgg19}) and DenseNet121 (Fig. \ref{fig:noise_ablation_dn121}) by $1.2$ and $8.2$ percentage points, respectively, while being outperformed by $0.8$ percentage points on ResNet18 (Fig. \ref{fig:noise_ablation_rn18}). The standard deviation of FlipOut at this point is lower than for noisy global magnitude for all networks tested, making it more robust to initial conditions and the noise sampling process. At this level, the addition of gradient noise to FlipOut also shows performance boosts compared to its non-noisy counterpart, namely $9.3$ percentage points for ResNet18, $3.2$ for VGG19 and $3.7$ for DenseNet121. The benefits caused by adding noise to global magnitude as compared to adding it to FlipOut are similar for VGG19; however, it is relatively small for ResNet18 at $2.6$ percentage points and even causes a $2$ percentage point drop in performance for DenseNet121.

Since FlipOut with $\lambda=1$ outperforms noisy global magnitude in 2 out of 3 cases for the highest level of sparsity while maintaining similar performance in all other cases as well as being less sensitive to the choice of seed, we conclude that its results cannot be explained only by the addition of noise and is also caused by the sign flips being taken into account when computing saliency.

Additionally, we conjecture that occurrences of under-shooting are indeed converted into over-shooting when adding gradient noise, allowing FlipOut to more accurately compute saliencies. This is evidenced by the fact that gradient noise addition benefits FlipOut more so than it does global magnitude, and implies that our method of dealing with deceitful shots is sound.

\section{Discussion}
\label{section:discussion}

In this work, we introduce the aim test, a general method for determining whether a point represents a local optimum for a weight during training, and propose using it for pruning by applying the test for all weights simultaneously and framing it as a saliency criterion. This method, coined FlipOut, demonstrates several desirable qualities: it is computationally tractable, allows for an exact level of sparsity to be selected, requires a single training run and has default hyperparameter settings which generate near optimal results, easing the burden of hyperparameter search. 

We compare the performance of FlipOut to relevant baselines from literature on a variety of object classification architectures. We show that it achieves state-of-the-art performance at the highest levels of sparsity tested for $2$ out of $3$ networks, and maintains similar performance in all other cases. Finally, we conduct an ablation study on the two components of our algorithm, gradient noise addition and the saliency criterion, and find that both play an important role in yielding its performance.
%
%
\bibliographystyle{splncs04}
\bibliography{main}

\end{document}